\documentclass[10pt,twocolumn,letterpaper]{article}

\usepackage{iccv}
\usepackage{times}
\usepackage{epsfig}
\usepackage{graphicx}
\usepackage{amsmath}
\usepackage{amssymb}

\usepackage{gensymb}
\usepackage{csquotes}
\usepackage{booktabs}
\usepackage{color}
\usepackage{amsmath}
\usepackage{amssymb}
\usepackage{multirow, varwidth}
\usepackage{verbatim}
\makeatletter
\newif\if@restonecol
\makeatother

\usepackage[ruled,vlined]{algorithm2e}
\usepackage{xr}
\usepackage[amsmath,thmmarks]{ntheorem}

\theorembodyfont{\normalfont}
\theoremseparator{.}

\theoremstyle{nonumberplain}

\newcommand{\argmax}{\operatornamewithlimits{argmax}}

\newcommand{\Comb}[2]{{}_{#1}C_{#2}}%

\DeclareRobustCommand\onedot{\futurelet\@let@token\@onedot}
\def\onedot{.\@\xspace}

\def\etal{\emph{et al}\onedot}

\usepackage{mathtools}

\usepackage[breaklinks=true,bookmarks=false]{hyperref}

\iccvfinalcopy 

\def\iccvPaperID{7671} 

\begin{document}

\title{ACAV100M: Automatic Curation of Large-Scale Datasets for \\Audio-Visual Video Representation Learning}

\author{
{\color{white}$\:\:\:\:\:\:$ }
Sangho Lee\footnotemark[1]\thanks{Equal Contribution},\;\;Jiwan Chung\footnotemark[1],\;\;Youngjae Yu,\;\;Gunhee Kim
{\color{white}$\:\:\:\:\:\:$ }\\
Seoul National University
\and
{\color{white}$\:$ }
Thomas Breuel,\;\;Gal Chechik
{\color{white}$\:$ }\\
NVIDIA Research
\and
{\color{white}$\:$ }
Yale Song 
{\color{white}$\:$ }
\\
Microsoft Research
\and
{\tt\href{https://acav100m.github.io}{\color{magenta}https://acav100m.github.io}}
}

\maketitle
\ificcvfinal\thispagestyle{empty}\fi

\begin{abstract}
The natural association between visual observations and their corresponding sound provides powerful self-supervisory signals for learning video representations, which makes the ever-growing amount of online videos an attractive source of training data. However, large portions of online videos contain irrelevant audio-visual signals because of edited/overdubbed audio, and models trained on such uncurated videos have shown to learn suboptimal representations. Therefore, existing approaches rely almost exclusively on datasets with predetermined taxonomies of semantic concepts, where there is a high chance of audio-visual correspondence. Unfortunately, constructing such datasets require labor intensive manual annotation and/or verification, which severely limits the utility of online videos for large-scale learning. In this work, we present an automatic dataset curation approach based on subset optimization where the objective is to maximize the mutual information between audio and visual channels in videos. We demonstrate that our approach finds videos with high audio-visual correspondence and show that self-supervised models trained on our data achieve competitive performances compared to models trained on existing manually curated datasets. The most significant benefit of our approach is scalability: We release ACAV100M that contains 100 million videos with high audio-visual correspondence, ideal for self-supervised video representation learning.
\end{abstract}

\section{Introduction}
\label{sec:introduction}

Our long-term objective is learning to recognize objects, actions, and sound in videos without the need for manual ground-truth labels.
This is not  only a theoretically interesting problem, since it mimics the development of auditory and visual perception by infants~\cite{gibson1969principles}, it is also of immense practical importance, since accurate manual labeling of audio-visual data is impractical.
Compared to self-supervised learning on static images~\cite{oord2018representation, hjelm2018learning, he2020momentum, chen2020simple}, audio-visual inputs pose additional challenges: large portions of a video may contain no relevant information, and auditory and visual inputs may not always be in correspondence.
Consequently, existing self-supervised methods on audio-visual data either start with datasets for which there is a high probability of audio-visual correspondence, or they learn audio-visual properties corresponding only to short-term statistical regularities.
The necessary datasets are usually manually created or rely on domain-specific properties
(e.g., \cite{carreira2019kinetics700,gemmeke2017audio} and below).
If we want to carry out self-supervised learning on full length (minutes, hours) of video without manually generating and/or selecting video clips, we need automated ways of curating such collections of audio/video clips from diverse collections of full length video.

\begin{figure}[t!]
    \begin{center}
        \includegraphics[trim=0.0cm 0cm 0cm 0.0cm,width=1.0\columnwidth]{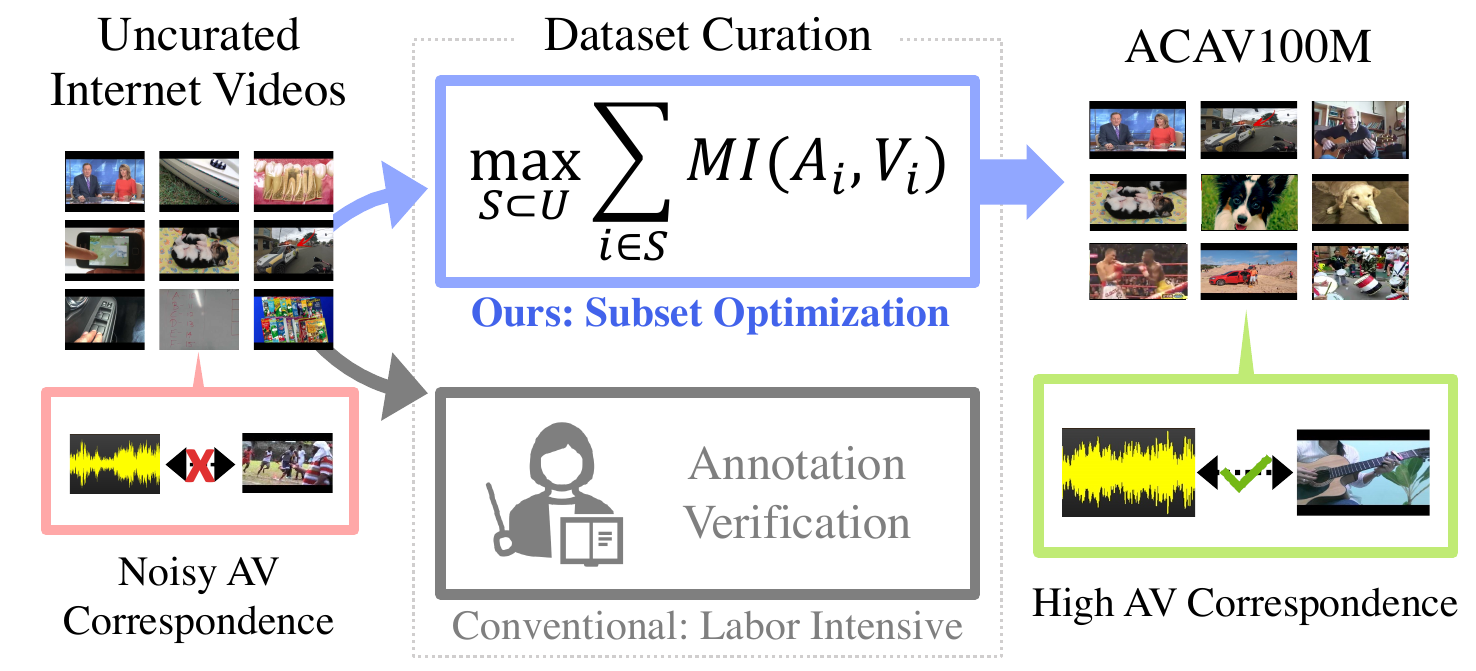}
    \end{center}
    \caption{We address the challenge of constructing a large-scale audio-visual dataset from uncurated Internet videos without relying on manual annotation or verification. We solve a constrained optimization problem that finds a subset maximizing the mutual information between audio and visual signals in videos. The result is a new 100M video dataset with high audio-visual correspondence, ideal for self-supervised video representation learning.}
    \vspace{-1em}
    \label{fig:keyidea}
\end{figure}

We consider self-supervised learning from unlabeled videos as a two-step process: (1) an automatic dataset curation process that generates short, relevant clips with useful self-supervisory signals, e.g., audio-visual correspondence, and (2) a self-supervised learning approach that operates on the collection of short clips. 
This paper focuses on step (1) and \textit{not} on step (2), providing an automated way of taking a collection of general or domain-specific videos of arbitrary length and reducing it to a collection of shorter clips containing a high portion of relevant audio-video correspondences.
The output of this step is a dataset, which can be used as input to existing self-supervised algorithms on audio-visual data~\cite{korbar2018cooperative, alwassel2019self, patrick2020multi}, as well as the development of novel self-supervised techniques.

To achieve step (1), we assume access to a large collection of unconstrained videos and solve a subset selection problem with an information-theoretic measure of audio-visual correspondence as a selection criterion. Specifically, we find a subset that maximizes mutual information (MI) between audio and visual channels of videos. This is a necessary condition for self-supervised learning approaches that rely on audio-visual correspondence~\cite{fisher2002probabalistic}. The main technical challenge we address is how to \textit{efficiently} measure the audio-visual MI and find a subset that maximizes the MI in a scalable manner. Given that video processing is notoriously compute and storage intensive, we put a particular emphasis on scalability, i.e., we want an approach that can easily handle hundreds of millions of video clips. 

MI estimation has a long history of research~\cite{paninski2003estimation, kraskov2004estimating}, including the recent self-supervised approaches~\cite{oord2018representation, hjelm2018learning, chen2020simple} that use noise contrastive estimation~\cite{gutmann2010noise} as the learning objective. While it is tempting to use such approaches to estimate MI in our work, we quickly encounter the ``chicken-and-egg'' problem: to obtain such models for estimating audio-visual MI, we need a training dataset where we can reliably construct positive pairs with a high probability of audio-visual correspondence; but that is what we are set out to find in the first place! One might think that randomly chosen videos from the Internet could be sufficient, but this has shown to produce suboptimal representations~\cite{alwassel2019self}; our empirical results also show that self-supervised models indeed suffer from noisy real-world audio-visual correspondences.

In this work, we turn to a clustering-based solution that estimates the MI by measuring the agreement between two partitions of data~\cite{meilua2007comparing, vinh2010information}. To circumvent the ``chicken-and-egg'' issue, we use off-the-shelf models as feature extractors and obtain multiple audio and visual clusters to estimate the MI. The use of off-the-shelf models is a standard practice in video dataset generation. Unlike existing approaches that use them as concept classifiers~\cite{heilbron2015activitynet, abu2016youtube, merler2019sports, nagrani2019voxceleb, chen2020vggsound}, here we use them as generic feature extractors. To avoid estimating the MI based on a restricted set of concepts the off-the-shelf models are trained on, we perform clustering over features computed across multiple layers (instead of just the penultimate layers), which has been shown to provide general feature descriptors not tied to specific concepts~\cite{yosinski2014transferable}.

To make our approach scalable, we avoid using memory-heavy components such as the Lloyd's algorithm~\cite{lloyd1982least} and instead use SGD~\cite{bottou1995convergence} to perform K-means clustering. Further, we approximately solve the subset maximization objective with a mini-batch greedy method~\cite{chen2013near}. Through controlled experiments with ground-truth and noisy real-world correspondences, we show that our clustering-based approach is more robust to the real-world correspondence patterns, leading to superior empirical performances than the contrastive MI estimation approaches. 

We demonstrate our approach on a large collection of videos at an unprecedented scale: We process 140 million full-length videos (total duration 1,030 years) and produce a dataset of 100 million 10-second clips (31 years) with high audio-visual correspondence. We call this dataset ACAV100M (short for \textbf{\underline{a}}utomatically \textbf{\underline{c}}urated \textbf{\underline{a}}udio-\textbf{\underline{v}}isual dataset of \textbf{\underline{100M}} videos). It is two orders of magnitude larger than the current largest video dataset used in the audio-visual learning literature, i.e., AudioSet~\cite{gemmeke2017audio} (8 months), and twice as large as the largest video dataset in the literature, i.e., HowTo100M~\cite{miech2019howto100m} (15 years).

To evaluate the utility of our approach in self-supervised audio-visual representation learning, we produce datasets at varying scales and compare them with existing datasets of similar sizes that are frequently used in the audio-visual learning literature, i.e., Kinetics-Sounds~\cite{arandjelovic2017look} at 20K-scale, VGG-Sound~\cite{chen2020vggsound} at 200K-scale, and AudioSet~\cite{gemmeke2017audio} at 2M-scale. Under the linear evaluation protocol with three downstream datasets, UCF101~\cite{soomro2012ucf101}, ESC-50~\cite{piczak2015dataset}, and Kinetics-Sounds~\cite{arandjelovic2017look}, we demonstrate that models pretrained on our datasets perform competitively or better than the ones pretrained on the baseline datasets, which were constructed with careful annotation or manual verification.

To summarize, our main contributions are: 1) We propose an information-theoretic subset optimization approach to finding a large-scale video dataset with a high portion of relevant audio-visual correspondences. 2) We evaluate different components of our pipeline via controlled experiments using both the ground-truth and the noisy real-world correspondence patterns. 3) We release ACAV100M, a large-scale open-domain dataset of 100M videos for future research in audio-visual representation learning.

\section{Related Work}
\label{sec:related_work}

\textbf{Large-Scale Data Curation}.
Several different types of audio-visual video datasets have been collected: 
(1) manually labeled, e.g., AudioSet~\cite{gemmeke2017audio}, AVE~\cite{tian2018audio}, 
(2) domain specific, e.g., AVA ActiveSpeaker~\cite{roth2019avaactivespeaker}, AVA Speech~\cite{chaudhuri2018avaspeech}, Greatest Hits~\cite{owens2016visually}, FAIR-Play~\cite{gao2019visualsound}, YouTube-ASMR-300K~\cite{yang2020telling},
and (3) unlabeled, unrestricted collections from consumer video sites, e.g., Flickr-SoundNet~\cite{aytar2016soundnet, arandjelovic2017look}.

{\it AudioSet}~\cite{gemmeke2017audio} contains about 2M clips corresponding to audio events retrieved from YouTube by keyword search; human raters verified the presence of audio events in the candidate videos.
{\it Moments in Time}~\cite{monfort2019moments} contains over one million clips of diverse visual and auditory events; video clips were selected using keywords (verbs) and manually reviewed for high correspondence between the clips and the keywords.
{\it HowTo100M}~\cite{miech2019howto100m} contains 136M clips segmented from 1.22M narrated instructional web videos retrieved by text search from YouTube, with an additional filtering step based on metadata.
{\it Web Videos and Text} (WVT)~\cite{stroud2020learning} contains 70M clips obtained by searching the web with keywords based on the Kinetics-700~\cite{carreira2019kinetics700} categories and retaining both the video and the associated text.
Chen \etal~\cite{chen2020vggsound} created a dataset of 200K clips for audio-visual research; clips were originally obtained by keyword search on YouTube and frames were classified with pretrained visual classifiers. Since keywords and visual classes do not perfectly correspond, such correspondences needed to be manually reviewed and corrected on randomly sampled clips in an iterative and interactive process.

We are building systems for learning audio-visual correspondence on diverse, unrestricted inputs. This requires large amounts of training data, making manual collection and labeling costly and impractical. Unlike previous dataset curation processes that involve costly human intervention, we introduce an automatic and scalable data curation pipeline for large-scale audio-visual datasets.

\textbf{Subset Selection}.
Our work focuses on data subset selection; extensive prior work exists in supervised~\cite{tsang2005core,wei2013using,shinohara2014submodular,wei2014submodular}, unsupervised~\cite{har2004coresets,wei2014unsupervised}, and active learning settings~\cite{lewis1994sequential,settles1995active}.
Different criteria for subset selection have been explored in the literature.
{\it Submodular functions} naturally model notions of information, diversity and coverage~\cite{wei2015submodularity}, and can be optimized efficiently using greedy algorithms~\cite{minoux1978accelerated,nemhauser1978submodular}.
{\it Geometric criteria} like the coreset~\cite{agarwal2005geometric} aim to approximate geometric extent measures over a large dataset with a relatively small subset.

Mutual-information (MI) between input feature values and/or labels has been used successfully~\cite{guo2010active,li2013adaptive,sourati2016classification} as a probablistically motivated criterion.
We propose to use MI as an objective function for subset selection
and make the following two unique contributions:
First, we use MI to measure audio-visual correspondence within videos by formulating MI between the audio and visual features.
Second, we apply MI for the large-scale video dataset curation problem.
In case of clustering-based MI estimation, we demonstrate that optimizing MI objective with a greedy algorithm is a practical solution for building a large-scale pipeline.

\section{Data Collection Pipeline}
\label{sec:data_collection}

Our pipeline consists of four steps:
(i) acquiring raw videos from the web and filtering them based on metadata, 
(ii) segmenting the videos into clips and extracting features with pretrained extractors,
(iii) estimating mutual information (MI) between audio and visual representations,
and (iv) selecting a subset of clips that maximizes the MI.

\subsection{Obtaining Candidate Videos}
\label{ssec:obtaining_candidate}

We crawl YouTube to download videos with a wide variety of topics. Unlike previous work that use a carefully curated set of keywords~\cite{chen2020vggsound}, which could inadvertently introduce bias, we aim for capturing the natural distribution of topics present in the website. To ensure the diversity in topics, cultures and languages, we create combinations of search queries with diverse sets of keywords, locations, events, categories, etc., to obtain an initial video list. 

Before downloading videos, we process the search results using metadata (provided by YouTube API) to filter out potentially low quality / low audio-visual correspondence videos. We use the duration to exclude videos shorter than 30 seconds (to avoid low quality videos) and longer than 600 seconds (to avoid large storage costs). We also exclude videos that contain selected keywords (in either title or description) or from certain categories -- i.e., gaming, animation, screencast, and music videos -- because most videos exhibit non-natural scenes (computer graphics) and/or low audio-visual correspondence. Finally, we detect language from the titles and descriptions using fastText~\cite{joulin2016fasttext,joulin2017bag} and keep the ones that constitute a cumulative ratio of $0.9$, resulting in eight languages (English, Spanish, Portuguese, Russian, Japanese, French, German, and Korean). 

The result is 140 million full-length videos with a total duration of 1,030 years (median: 198 seconds). To minimize the storage cost we download 360p resolution videos; this still consumes 1.8 petabytes of storage. Handling such large-scale data requires a carefully designed data pipeline. We discuss our modularized pipeline below.

\subsection{Segmentation \& Feature Extraction}
\label{ssec:segmentation_feature}

\textbf{Clip Segmentation}.
To avoid redundant clips, we extract up to three 10-second clips from each full-length video. We do this by detecting shot boundaries (using the \texttt{scdet} filter in FFmpeg) and computing pairwise clip similarities based on the MPEG-7 video signatures (using the \texttt{signature} filter in FFmpeg). We then select up to 3 clips that give the minimum total pairwise scores using local search~\cite{johnson1988localsearch}. This gives us about 300M clips.

\textbf{Feature Extraction}.
To measure correspondence  between audio and visual channels of the 300M clips, we need good feature representations. An ideal representation would capture a variety of important aspects from low-level details (e.g., texture and flow) to high-level concepts (e.g., semantic categories). However, such an oracle extractor is hard to obtain, and the sheer scale of data makes it impractical to learn optimal feature extractors end-to-end. Therefore, we use the ``off-the-shelf'' pretrained models to extract features, i.e., SlowFast~\cite{feichtenhofer2019slowfast} pretrained on Kinetics-400~\cite{kay2017kinetics} and VGGish~\cite{hershey2017cnn} pretrained on YouTube-8M~\cite{abu2016youtube} for visual and audio features, respectively.

\subsection{Subset Selection via MI Maximization}
\label{sec:mie_ss}
Next, we select clips that exhibit strong correspondence between visual and audio channels. To this end, we estimate the mutual information (MI) between audio and visual signals. Computing the exact MI is infeasible because it requires estimating the joint distribution of high dimensional variables, but several approximate solutions do exist~\cite{walters2009estimation}. Here we implement and compare two approaches: a noise-contrastive estimator (NCE)~\cite{gutmann2010noise}, which measures MI in a continuous feature space, and a clustering-based estimator that computes MI in a discrete space via vector quantization. The former estimates MI for each video clip, while the latter estimates MI for a set of video clips. As we show later in our experiments, we find the clustering-based MI estimator to be more robust to real-world noise.

\subsubsection{NCE-based MI Estimation}
\label{ssec:selection_nce}
Contrastive approaches have become a popular way of estimating MI between different views of the data~\cite{oord2018representation,hjelm2018learning}. We add linear projection heads over the precomputed audio/visual features and train them using the contrastive loss~\cite{chen2020simple}. From a mini-batch $\{(v_i, a_i)\}_{i=1}^{N_b}$ where $v_i$ and $a_i$ are visual and audio features, respectively, we minimize
\begin{equation}
    \label{eq:contrastive}
    l(v_i,{a_i)} = -\log \frac{\exp(S(\mathbf{z}_i^v, \mathbf{z}_{i}^a) / \tau)}{\sum_{j=1}^{N_b} \exp(S(\mathbf{z}_i^v, \mathbf{z}_{j}^a) / \tau)},
\end{equation}
\noindent
where $\mathbf{z}_i^v$ and $\mathbf{z}_i^a$ are embeddings from the linear projection heads, $S(\cdot, \cdot)$ measures the cosine similarity, and $\tau$ is a temperature term (we set $\tau=0.1$). For each mini-batch we compute $l(v_i, a_i)$ and $l(a_i, v_i)$ to make the loss symmetric. 

Once trained, we can directly use $S(\mathbf{z}^v, \mathbf{z}^a)$ to estimate audio-visual MI and find a subset by taking the top $N$ candidates from a ranked list of video clips.

\subsubsection{Clustering-based MI Estimation}
\label{ssec:selection_clustering}

\textbf{MI Estimation.} Clustering is one of the classical ways of estimating MI~\cite{meilua2007comparing,vinh2010information}. Given two partitions of a dataset $\mathbf{X}$ w.r.t. audio and visual features, $\mathcal{A} = \{\mathbf{A}_1, \cdots, \mathbf{A}_{|\mathcal{A}|}\}$ and $\mathcal{V} = \{\mathbf{V}_1, \cdots, \mathbf{V}_{|\mathcal{V}|}\}$, we estimate their MI as:
\begin{equation}
    \label{eq:clustering_mi}
    \mbox{MI}(\mathcal{A}, \mathcal{V}) = \sum_{i=1}^{|\mathcal{A}|} \sum_{j=1}^{|\mathcal{V}|} \frac{|\mathbf{A}_i \cap \mathbf{V}_j|}{|\mathbf{X}|} \log \frac{|\mathbf{X}||\mathbf{A}_i \cap \mathbf{V}_j|}{|\mathbf{A}_i||\mathbf{V}_j|}.
\end{equation}
\noindent
This formulation estimates MI in a discrete (vector-quantized) space induced by clustering, and thus the quality of clustering affects the quality of the estimator. A straightforward approach to obtaining $\mathcal{A}$ and $\mathcal{V}$ is to cluster videos using the output from the penultimate layers of the pretrained networks. However, this can introduce distributional bias specific to the datasets on which the networks are pretrained~\cite{yosinski2014transferable,wang2018deep}. To address this issue, we cluster samples over each output space induced by different layers of the networks. This allows the MI estimator to consider a wide range of abstract concepts, from low-level (such as textures) to high-level (such as object parts)~\cite{bau2019dissection}.  

Specifically, we use the feature spaces induced by the five convolutional blocks from each of the SlowFast and VGGish feature extractors. We then compute the average MI between \textit{all pairs} of clusterings as our MI estimator. Let $\mathcal{CV}_{\mathbf{X}}^{(i)} = \{\mathbf{V}_1^{(i)}, \cdots, \mathbf{V}_{n_i}^{(i)}\}$ and $\mathcal{CA}_{\mathbf{X}}^{(i)} = \{\mathbf{A}_1^{(i)}, \cdots, \mathbf{A}_{m_i}^{(i)}\}$ denote the clustering results induced by the $i$-th convolutional block of the visual and audio feature extractors, respectively. We compute:
\begin{equation}
    \label{eq:clustering_mi2}
    F(\mathbf{X}) = 
    \sum_{(\mathcal{X}, \mathcal{Y}) \in \mathcal{C}_\mathbf{X}} \frac{\mbox{MI} (\mathcal{X}, \mathcal{Y})}{{\Comb{10}{2}}},
\end{equation}
\noindent
where $\mathcal{C}_\mathbf{X}$ denotes the combination of two elements from $\{\mathcal{CV}_\mathbf{X}^{(i)}\}_{i=1}^{5} \cup \{\mathcal{CA}_\mathbf{X}^{(j)}\}_{j=1}^{5}$ and $\Comb{10}{2}$ denotes the number of 2-combinations out of 10 elements, which equals to 45.
This computes MI between layers from both within and across the extractors of different modalities (referred to as \textit{combination} pairing scheme in Section~\ref{ssec:correspondence_results}).

\begin{algorithm}[tp]
\SetAlgoNoLine
\DontPrintSemicolon
\textbf{Input:} initial dataset $\mathbf{D}$, MI estimator $F$, target subset size $M$, batch size $b$, selection size $s$\\
\textbf{Output:} $\mathbf{X} \subseteq \mathbf{D}, |\mathbf{X}| = M$\\
$\mathbf{X}_0 \gets \emptyset, i \gets 0$\\
\While{$|X_i| < M$}{
    Randomly sample $\mathbf{B} \subseteq \mathbf{D} \backslash \mathbf{X}_{i}, |\mathbf{B}| = b$\\
    $\mathbf{Y}_0 \gets \emptyset, j \gets 0$\\
    \While{$j < s$}{
        $x \gets \argmax_{x \in \mathbf{B} \backslash \mathbf{Y}_j} F(\mathbf{X}_i \cup \mathbf{Y}_{j} \cup \{x\})$\\
        $\mathbf{Y}_{j+1} \gets \mathbf{Y}_{j} \cup \{x\}, j \gets j + 1$\\
        \lIf{$|\mathbf{X}_i \cup \mathbf{Y}_j| = M$}{
            break
        }
    }
    $\mathbf{X}_{i+1} \gets \mathbf{X}_i \cup \mathbf{Y}_j, i \gets i + 1$
}
$\mathbf{X} \gets \mathbf{X}_i$\\
\textbf{Return} $\mathbf{X}$
\caption{Batch Greedy Subset Selection}
\label{alg:batch_greedy}
\end{algorithm}
\noindent

\textbf{Batch Greedy Subset Selection.}
Since the MI estimator $F(\cdot)$ is a function of $\mathbf{X}$, we can formulate an optimization problem where the goal is to find a subset $\mathbf{X}$ that maximizes $F(\mathbf{X})$. In general, finding a global solution to problems such as ours is NP-hard and thus greedy heuristic solutions are used instead~\cite{nemhauser1978analysis}. However, they typically select one sample in each iteration and re-evaluate the goodness function, e.g., $F(\cdot)$, on all the remaining candidates. This introduces a challenge to our setting because the time complexity is quadratic to the size of the population; this is clearly not scalable to 300 million instances. 

Therefore, we approximate the typical greedy solution using the batch greedy algorithm~\cite{chen2013near}, as shown in Algorithm~\ref{alg:batch_greedy}. It randomly samples a batch $\mathbf{B}$ from the remaining pool of candidates, and searches for the next element to be included in the active solution set only within $\mathbf{B}$. This batch trick reduces the time complexity down to linear, i.e., $O(N \times |\mathbf{B}|)$, where $N$ is the size of the input dataset. We demonstrate the efficacy of the algorithm in Section~\ref{sec:correspondence}.

\textbf{Stochastic Clustering}.
One missing piece in this pipeline is an \textit{efficient} clustering algorithm scalable to hundreds of millions of instances. The most popular choice among various clustering methods is K-means clustering~\cite{wu2008top}, which is a special case of mixture density estimation for isotropic normal and other densities. Typically, an expectation-maximization (EM) algorithm, such as Lloyd's~\cite{lloyd1982least}, is used to find the cluster centers. Such algorithms require repeated computation of the distances of all samples from all $k$ cluster centers, followed by cluster assignment, until convergence. Lloyd's algorithm updates cluster centers only after each pass through the entire dataset. But for very large datasets (like ours), a small subset usually contains enough information to obtain good estimates of the cluster centers, meaning that EM-style algorithms tend to take (perhaps too) many epochs to converge.

There are different strategies for addressing this issue, including random sampling and subsetting, but a straightforward approach is to replace EM algorithm with an SGD~\cite{martinetz1991gas, bottou1995convergence, sculley2010web}. In such an approach, for large datasets, convergence rate and final accuracy of the cluster centers are determined not by the total dataset size, but by the learning rate schedule.
A straightforward SGD update rule is to compute the nearest cluster centers for each sample in a batch and then update the cluster centers using a convex combination of the cluster centers and their nearest samples, weighting the samples with a learning rate $\lambda$ and the cluster centers with $(1-\lambda)$.
However, mixture density estimators in general suffer from the problem that adding mixture components with zero probability does not change the mixture density; in practice, this means EM and SGD-based algorithms may end up with cluster centers that stop receiving updates at some point during the optimization. 

We address this problem by estimating the mixture component utilization rate as the ratio of the total number of updates to the cluster center divided by the total number of estimation steps, and reinitializing cluster centers when that probability falls below $(1/k)^2$. In Section~\ref{ssec:correspondence_results}, we demonstrate that our mini-batch SGD update shows comparable accuracy to batch update in correspondence retrieval tasks.

\begin{table*}[tp]
\centering
\small
\begin{tabular}[t]{l||c|c||c|c||c}
~ & \multicolumn{2}{c||}{Natural Class Correspondence} & \multicolumn{2}{c||}{Arbitrary Class Correspondence} & Audio-Visual \\
Method & CIFAR10-Rotation & CIFAR10-Flip & MNIST-CIFAR10 & MNIST-FSDD & Kinetics-Sounds \\ \hline
Ranking-inner & 87.872 $\pm$ 0.002 & 87.044 $\pm$ 0.001 & 63.076 $\pm$ 0.001 & 64.453 $\pm$ 0.003 & 52.558 $\pm$ 0.002 \\
Ranking-cos & 87.872 $\pm$ 0.002 & 87.044 $\pm$ 0.001 & 67.600 $\pm$ 0.002 & 61.893 $\pm$ 0.004 & 60.108 $\pm$ 0.001 \\
Ranking-$l_2$ & 87.872 $\pm$ 0.002 & 87.044 $\pm$ 0.001 & 66.796 $\pm$ 0.001 & 62.933 $\pm$ 0.003 & 51.236 $\pm$ 0.001 \\ \hline
Ours-Contrastive & \textbf{99.395} $\pm$ 0.000 & \textbf{99.480} $\pm$ 0.001 & 73.252 $\pm$ 0.040 & \textbf{73.733} $\pm$ 0.027 & 73.066 $\pm$ 0.036 \\ 
Ours-Clustering & 87.292 $\pm$ 0.014 & 87.248 $\pm$ 0.010 & \textbf{77.224} $\pm$ 0.009 & 69.440 $\pm$ 0.049 & \textbf{88.705} $\pm$ 0.004 \\ \hline
\end{tabular}
\vspace{1em}
\caption{Correspondence retrieval results. We conduct a total of five runs and report the precision with the 99\% confidence interval. We use the clustering pairing scheme which gives the highest score in each configuration: combination, except diagonal for Ranking-inner, Ranking-cos and Rank-$l_2$ on CIFAR10-Rotation and CIFAR10-Flip.}
\vspace{-1em}
\label{tab:correspndence_results}
\end{table*}

\section{Evaluation on Correspondence Retrieval}
\label{sec:correspondence}

We systematically evaluate different components of our pipeline with synthetic correspondence-retrieval tasks, where we generate corresponding and non-corresponding pairs using CIFAR-10~\cite{krizhevsky2009learning}, MNIST~\cite{lecun1998gradient} and FSDD~\cite{zohar2018fsdd}.  
In each correspondence retrieval task, the goal is to discover the known corresponding samples among the  non-corresponding pairs. To show the generality of the findings, we also experiment with Kinetics-Sounds~\cite{arandjelovic2017look} which exhibit real-world audio-visual correspondence.

\subsection{Experimental Setting}
\label{ssec:correspondence_datasets}
\paragraph{Datasets} We construct five datasets where each instance is a pair of samples with different correspondence types.

\textbf{1/2) CIFAR10-Rotation/Flip}. 
We use images from five randomly selected categories to construct a ``positive pair'' set, and use the rest for a ``negative pair'' set. For the positive set, we create pairs of images by sampling two different images from the same category (e.g., two images of a bird), and apply a geometric transformation to one of them; we apply either a 90\degree~CCW rotation (CIFAR10-Rotation) or a horizontal flip (CIFAR10-Flip). The negative set follows the same process but each pair contains images from different categories. We categorize this type of correspondence as ``Natural Class Correspondence'' because pairings are made over natural semantic categories. 

\textbf{3/4) MNIST-CIFAR10/FSDD}.
We use images from five digit categories to construct a positive set and use the rest for a negative set. Different from above, correspondence is defined via an arbitrary class-level mapping, e.g., ``digit 0'' images map to the ``car'' images in CIFAR-10 or ``digit 0'' audio samples in FSDD. We take samples from the same categories to construct the positive set and samples from different categories for the negative set. We call these ``Arbitrary Class Correspondence'' to differentiate from above.

\textbf{5) Kinetics-Sounds}. 
Unlike the above datasets where the correspondence is defined over class categories, here the correspondence is defined at the \textit{sample} level, i.e., a positive set contains pairs of audio and visual channels of the same video, and a negative set contains randomly permuted pairs. We do not utilize class labels to construct the dataset.

\vspace{-1em}\paragraph{Methods}
We compare our pipeline (both contrastive-based and clustering-based) to three ranking-based approaches. All the methods use the same precomputed features. For images, we use ResNet-50~\cite{he2016deep} pretrained on ImageNet~\cite{deng2009imagenet}. For videos, we use SlowFast~\cite{feichtenhofer2019slowfast} pretrained on Kinetics-400~\cite{kay2017kinetics} and VGGish~\cite{hershey2017cnn} pretrained on YouTube-8M~\cite{abu2016youtube} for visual and audio features, respectively. For the ranking baselines, we apply PCA~\cite{pearson1901liii} to reduce the feature dimensionality to 64 and rank the instances based on three similarity metrics: inner product, cosine similarity, and (negative) $l_2$ distance. Because all our datasets have an equal number of positive and negative instances, we simply select the top 50\% instances as the retrieval result. 

\vspace{-1em}\paragraph{Protocol} 
We split each dataset into train and test partitions of the same size. We conduct a total of five runs for each of the five datasets and report results on the test splits. We use train sets only for the contrastive estimator to train the projection heads. When constructing each dataset, we sample at most $n=1000$ instances from each category of the source datasets. For the noise contrastive estimator, we train the linear projection heads for 100 epochs using the AMSGrad of Adam optimizer~\cite{reddi2018convergence} with a learning rate of 2e-4. We randomly take one sample from each class to build a mini-batch for class-level correspondence datasets, and sample random $N_b=10$ clips to build a mini-batch for the sample-level correspondence dataset. When applying our clustering-based method, we perform the SGD K-means clustering with the ``ground-truth'' number of centroids as the number of classes in each source dataset; we use the batch greedy algorithm with a batch size $b=100$ and a selection size $s=25$.

\subsection{Ablation Results \& Discussion}
\label{ssec:correspondence_results}

Table~\ref{tab:correspndence_results} shows that the two variants of our approach -- contrastive and clustering -- achieve overall higher precision rates than the ranking baselines. The contrastive approach performs well on the two datasets with the ``natural class correspondence,'' conforming to the previous results that shows contrastive learning is robust to geometric transformations~\cite{chen2020simple}. The clustering approach excels on Kinetics-Sounds that contains natural audio-visual correspondence, which is closer to our intended scenario. Therefore, we conduct various ablation studies on Kinetics-Sounds to validate different components of our clustering-based approach.

\begin{table}[t]
\centering
\small
\begin{tabular}[t]{l|c|c}
Layers & Method & Precision \\ \hline
\multirow{5}{*}{Single} & Layer1 & 50.820 $\pm$ 0.014 \\
 & Layer2 & 51.412 $\pm$ 0.011 \\
 & Layer3 & 52.659 $\pm$ 0.012 \\
 & Layer4 & 54.422 $\pm$ 0.012 \\
 & Layer5 & 58.418 $\pm$ 0.030 \\ \hline
\multirow{3}{*}{Multiple} & Diagonal & 71.450 $\pm$ 0.005 \\
 & Bipartite & 76.969 $\pm$ 0.005 \\
 & Combination & \textbf{88.705} $\pm$ 0.004 \\ \hline
\end{tabular}
\vspace{1em}
\caption{Correspondence retrieval results on Kinetics-Sounds with different clustering pairing schemes. We conduct a total of five runs and report the precision with the 99\% confidence interval.}
\label{tab:correspondence_similarity}
\end{table}

\textbf{Multi-Layer Clustering}.
All the feature extractors that we use consist of five convolutional blocks. As discussed in Section~\ref{ssec:selection_clustering}, we cluster samples over each of the five output spaces to capture a wide range of abstract concepts. This raises a question: How should we combine audio-visual clusters for MI estimation? Table~\ref{tab:correspondence_similarity} compares the single-layer approaches to multi-layer approaches. Each of the single-layer approach estimates the audio-visual MI based on a single pair of clustering results. We can see that the precision increases as we use clustering results from higher layers. However, all single-layer methods perform significantly worse than multi-layer variants.

We explore three options to select pairs of clusterings for MI estimation. \texttt{Diagonal} computes an average MI across all five single-layer scores (with $L$ layers, this computes MI $L$ times), \texttt{Bipartite} computes an average MI between all possible combinations of audio-visual clustering results ($L^2$ times), and \texttt{Combination} (\textbf{ours}) computes an average MI between all possible combinations of clustering results, regardless of modalities ($\Comb{2L}{2}$ times).
We observe that the performance increases with the number of connections as shown in the bottom rows of Table~\ref{tab:correspondence_similarity}. This positive relationship suggests that the consensus between layers from the same extractor, as well as that across extractors, contributes to the clarity of correspondence signal. We further experimented with different layer weights for the \texttt{Combination} approach and found it to be robust to different weight distributions; we provide the results in the supplementary material.

\begin{figure}[t]
   \centering
   \includegraphics[width=0.47\textwidth]{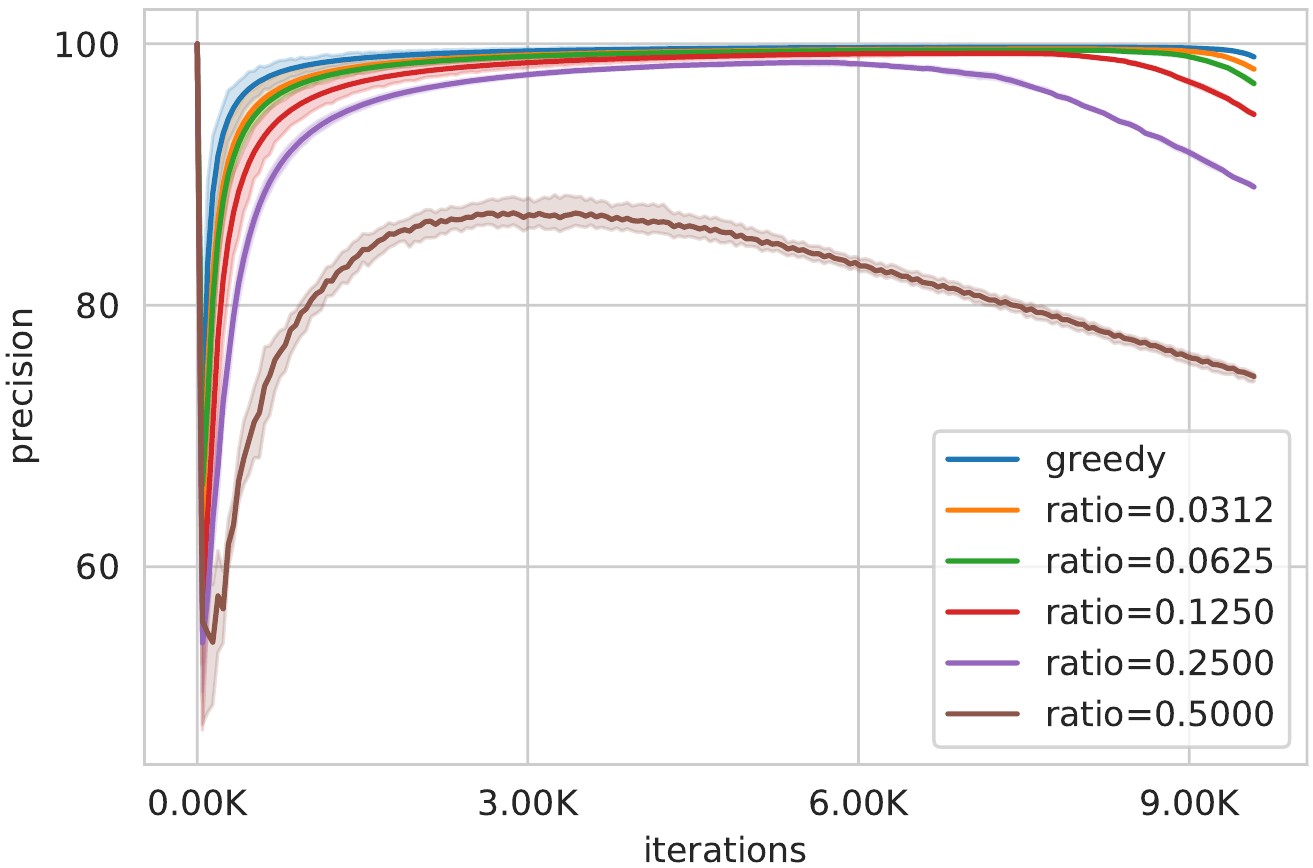} 
   \caption{\textbf{Greedy vs. batch greedy algorithms} with varying selection-to-batch size ratios, $s/b$. The shaded regions show 99\% confidence intervals obtained by five runs on Kinetics-Sounds. The batch greedy algorithm is robust when the ratio is $\leqslant$ 25\%.}
   \label{fig:correspondence_ratio}
\end{figure}

\begin{figure}[t]
   \centering
   \includegraphics[width=0.47\textwidth]{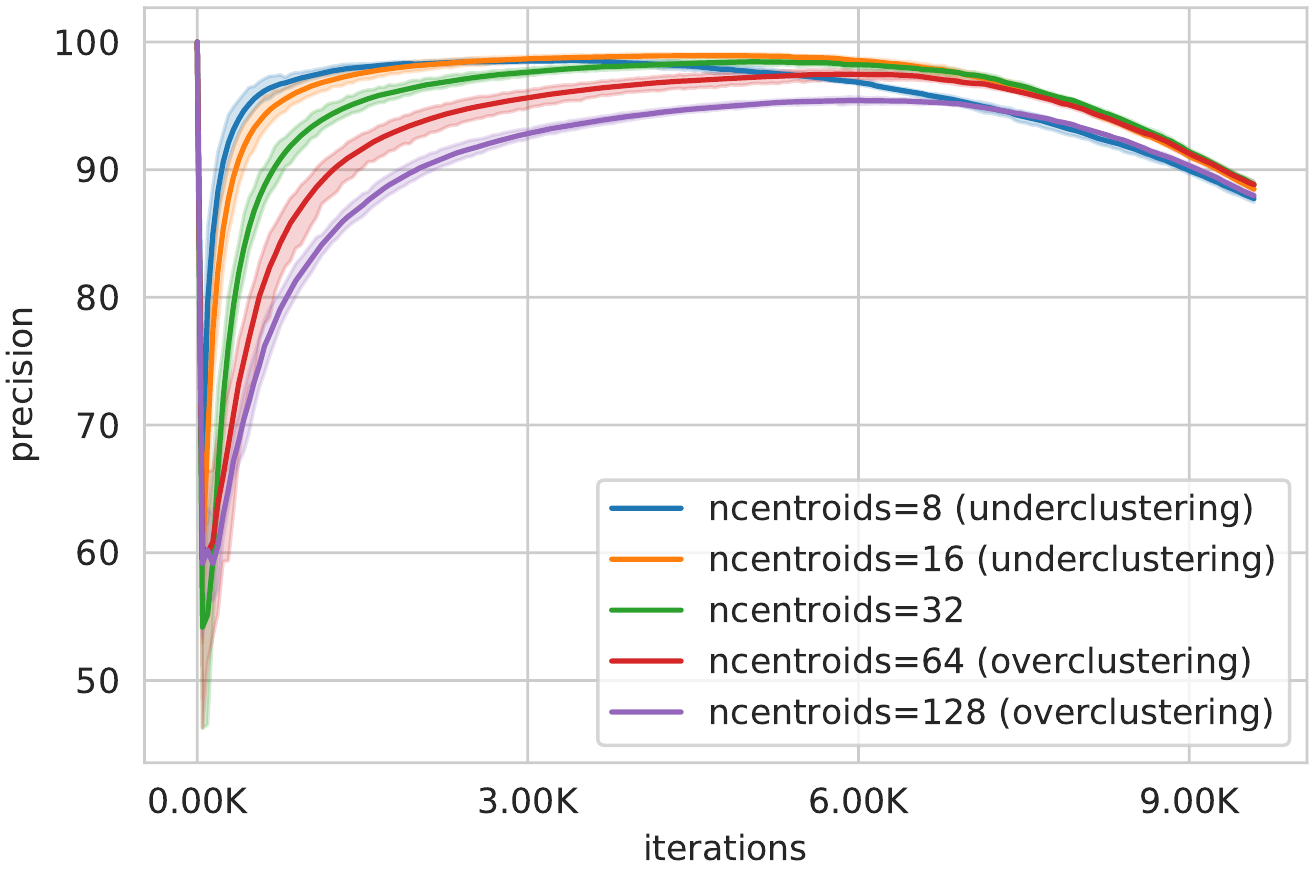} 
   \caption{\textbf{Sensitivity analysis on the number of centroids}. We determine under/over-clustering based on the ground-truth number of class categories in Kinetics-Sounds ($c=32$). The shaded regions show 99\% confidence intervals over five runs. }
   \label{fig:correspondence_ncentroids}
\end{figure}

\textbf{Mini-Batch SGD K-means Clustering}.
We compared mini-batch SGD K-means~\cite{bottou1995convergence} to the standard EM (Lloyd's) approach~\cite{lloyd1982least} and obtained very similar results on Kinetics-Sounds: 88.705 $\pm$ 0.004 (SGD) versus 88.732 $\pm$ 0.005 (EM). This shows that our SGD solution has negligible performance degradation while enjoying a significantly less memory requirement than the standard EM approach.

\textbf{Batch Greedy Subset Selection}.
We explore how the use of mini-batches affects the quality of the selected subsets. We compare the greedy algorithm and the batch greedy algorithm with a batch size $b=160$ and varying selection sizes $s=\{5, 10, 20, 40, 80\}$. As shown in Figure~\ref{fig:correspondence_ratio}, the performance gap between the greedy algorithm and the batch greedy algorithm is marginal (greedy: 98.970 vs. batch greedy with $(b, s) = (160, 5)$: 98.020), which validates our use of the batch greedy algorithm. While the batch size itself does not have a large impact on the subset quality, the ratio of selection size to batch size ($s/b$) highly affects the retrieval performance; the performance drops sharply as the ratio exceeds 0.25 in several ($b$, $s$) configurations.
This is mainly dataset-dependent: by construction, there is a 50\% chance that a sample will be a positive. We believe that the constructed dataset contains roughly 25\% \textit{easy positives}, i.e., videos with very high correspondence. When the selection ratio $s/b$ does not exceed the easy positive ratio, the batch greedy algorithm finds those videos without introducing false positives, providing robustness. We found similar patterns with other ratios of $s/b > 25\%$.

\textbf{Number of Centroids}.
We vary the number of centroids $k\in\{ 8,16,32,64,128 \}$ to see how sensitive our approach is to the parameter. We apply the batch greedy algorithm with a batch size $b=100$ and a selection size $s=25$ on Kinetics-Sounds. Figure~\ref{fig:correspondence_ncentroids} shows that, although the final performance is similar across different number of centroids, they show different trends: underclustering ($k=\{ 8, 16\}$) shows high precision in early iterations while overclustering ($k=\{ 64, 128 \}$) shows slower drop in the later stage.

\begin{figure*}[t!]
    \vspace{-1em}
    \begin{center}
        \includegraphics[trim=0.0cm 0cm 0cm 0.0cm,width=0.98\textwidth]{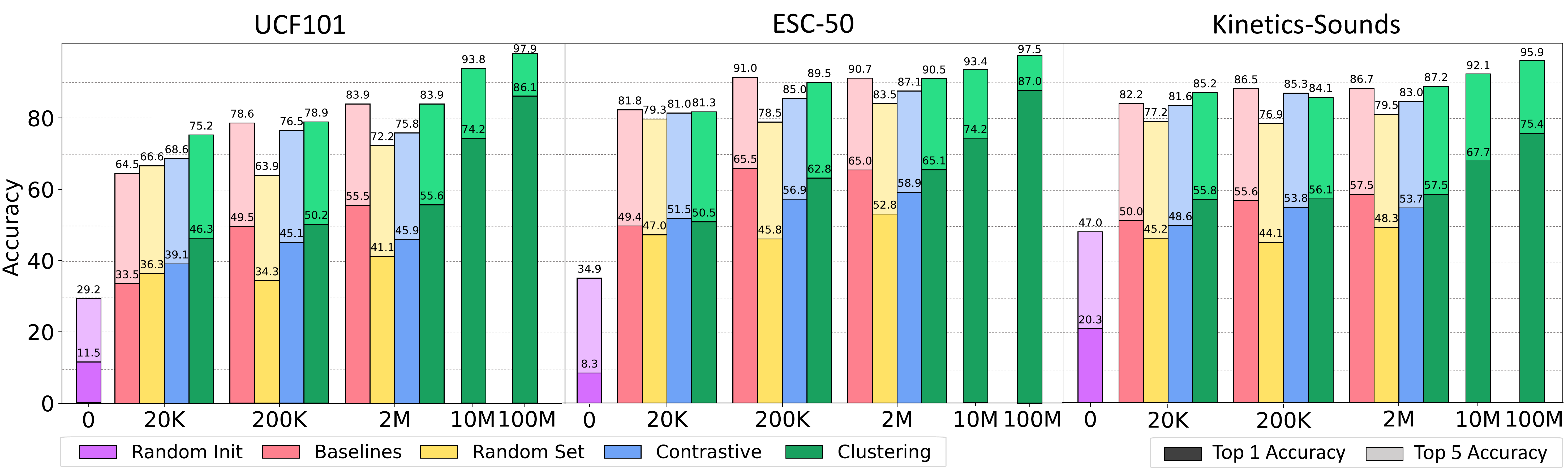}
    \end{center}
    \vspace{-1em}
    \caption{\textbf{Linear evaluation on downstream tasks}. The top-1/5 accuracy (\%) of video classification on UCF101~\cite{soomro2012ucf101}, audio classification on ESC-50~\cite{piczak2015dataset} and audio-visual classification on Kinetics-Sounds (KS)~\cite{arandjelovic2017look}. We group the results by the downstream tasks and by the scale of the pretrain datasets. Baselines are Kinetics-Sounds~\cite{arandjelovic2017look} (20K), VGG-Sound~\cite{chen2020vggsound} (200K), and AudioSet~\cite{gemmeke2017audio} (2M).}
    \label{fig:linear evaluation}
\end{figure*}

\section{Large-Scale Evaluation}
\label{sec:evaluation}

We construct datasets at varying scales (20K, 200K, 2M) and compare them to existing datasets often used in the audio-visual learning literature: Kinetics-Sounds~\cite{arandjelovic2017look} (20K), VGG-Sound~\cite{chen2020vggsound} (200K), and AudioSet~\cite{gemmeke2017audio} (2M). Note that all three datasets involve either human annotation~\cite{arandjelovic2017look,gemmeke2017audio} or manual verification~\cite{chen2020vggsound}. To demonstrate the scalable nature of our approach, we also generate datasets with 10M and 100M videos and evaluate their performance.

For the contrastive approach, we train linear projection heads on a batch size of 1024 from a randomly drawn set of 100M videos. Note that these additional videos are only used to train projection heads for MI estimation (Sec. 3.3.1), which is discarded once dataset curation is finished; all approaches use the same number of videos under the same evaluation protocol on all downstream tasks. We train the model for three epochs and rank the entire video set (300M) based on the cosine similarity~\cite{chen2020simple}. We then take top $N\in\{20\text{K}, 200\text{K}, 2\text{M}\}$ ranked videos for the final dataset. For the clustering-based variant, we vary the number of clusters $C\in\{100, 200, 500, 1000, 2000\}$ for each size of the datasets.

\subsection{Linear Evaluation on Downstream Tasks}
\label{ssec:downstream}
To assess the quality of the datasets, we pretrain identical models on different datasets and evaluate their performance on downstream tasks. The idea is that if a model performed particularly better than the others, the dataset used to train that model must be superior to the other datasets. We pretrain audio-visual CNNs from scratch using the self-supervised objective of SimCLR~\cite{chen2020simple}; we use 3D ResNet-50~\cite{christoph2016spatiotemporal} and ResNet-50~\cite{he2016deep} as the visual and audio CNNs, respectively. We follow the linear evaluation protocol~\cite{chen2020simple} by adding a linear classifier on top of the learned and frozen models. We test on three downstream tasks: visual action recognition on UCF101~\cite{soomro2012ucf101}, sound classification on ESC-50~\cite{piczak2015dataset}, and audio-visual action recognition on Kinetics-Sounds~\cite{arandjelovic2017look} (we concatenate audio-visual features for the linear classifier). Note that the training procedures are identical for all the models except for the datasets used to train them. We report mean accuracy across the official splits of UCF101 and ESC-50. We provide details of these experimental settings in the supplementary material. 

Figure~\ref{fig:linear evaluation} shows that models pretrained on our dataset (green bars) achieve similar, or even slightly better, performances compared to the baseline datasets (pink bars) at 20K, 200K, and 2M scales. The significant gap between ours vs. random set (yellow bars) shows the improvement does not come from the initial pool we crawl (the 300M set) but rather come from higher portion of audio-visual correspondence in the resulting dataset. Our clustering approach to MI estimation (green bars) generally outperforms the contrastive approach (blue bars), suggesting its robustness to noisy real-world audio-visual correspondences.  Finally, we report the results obtained from 10M and 100M datasets produced with our clustering-based MI estimation module (we omit the baseline results at these scales due to computational reasons). The significant performance boost from the 10M and 100M models reaffirms the importance of large-scale training. Considering our data curation process does not involve human intervention (i.e., no manual annotation and verification) this is a promising result showing the potential for large-scale self-supervised learning: one can obtain datasets of arbitrary scales and develop self-supervised models by leveraging high portion of audio-visual correspondences provided in the datasets.

\subsection{Human Evaluation}
\label{ssec:human_evaluation}
We conduct a user study to assess the perceived presence/absence of audio-visual correspondence in video clips. We compare clips from four datasets: AudioSet~\cite{gemmeke2017audio}, VGG-Sound~\cite{chen2020vggsound}, ours with clustering (2M scale, 1K clusters), and random (drawn from the 300M set). We prepare 100 randomly sampled clips from each of these datasets, for a total of 400 clips. We recruit 12 participants and present each with 100 clips (25 clips per dataset), and ask them whether audio and visual are corresponding or not. This provides us with 3 votes per video (we provide the details of the questionnaire in the supplementary material).

Table~\ref{tab:human_evaluation} shows the majority voting accuracy and inter-rater agreement (measured by Fleiss' Kappa~\cite{fleiss1971measuring}). Every dataset has Fleiss' Kappa greater than 0.4, verifying the reliability of the accuracy statistics~\cite{landis1977measurement}. Ours significantly improves audio-visual correspondence over a random subset (69\% vs. 44\%), and is even rated slightly higher than AudioSet. The annotation process for AudioSet has focused on audio events so we suspect that several of videos do not contain \textit{visible} sound sources. There is still a significant gap between ours and VGG-Sound; we note that our process finds audio-visual correspondence without relying on manual verification as was done in VGG-Sound.

\begin{table}[t]
\centering
\small
\begin{tabular}[t]{l|c|c}
Dataset & Majority Vote (\%) & Fleiss' Kappa \\ \hline
AudioSet & 65.66 & 0.4385 \\
VGG-Sound & 84.00 & 0.4634 \\
Ours (2M) & 69.00 & 0.5110 \\ 
Random & 44.00 & 0.6112 \\ \hline
\end{tabular}
\vspace{.5em}
\caption{\textbf{Human evaluation results} assessing the perceived audio-visual correspondence in videos from different datasets.}
\label{tab:human_evaluation}
\end{table}

\section{Conclusion}
\label{sec:conclusion}
This work complements existing line of research on self-supervised representation learning with three main contributions: i) proposing an automatic and scalable data collection pipeline for audio-visual representation learning, ii) demonstrating that the MI-based subset selection can retrieve correspondence in both artificial and practical settings, and iii) releasing a large-scale open-domain video dataset consisting of 100M clips curated with our pipeline.

\textbf{Acknowledgements.}
Authors in Seoul National University are supported by Institute of Information \& communications Technology Planning \& Evaluation (IITP) grant funded by the Korea government (MSIT) (No.2017-0-01772, Video Turing Test, No.2019-0-01082, SW StarLab).


\appendix

\begin{figure*}[!h]
    \centering
    \includegraphics[width=\textwidth]{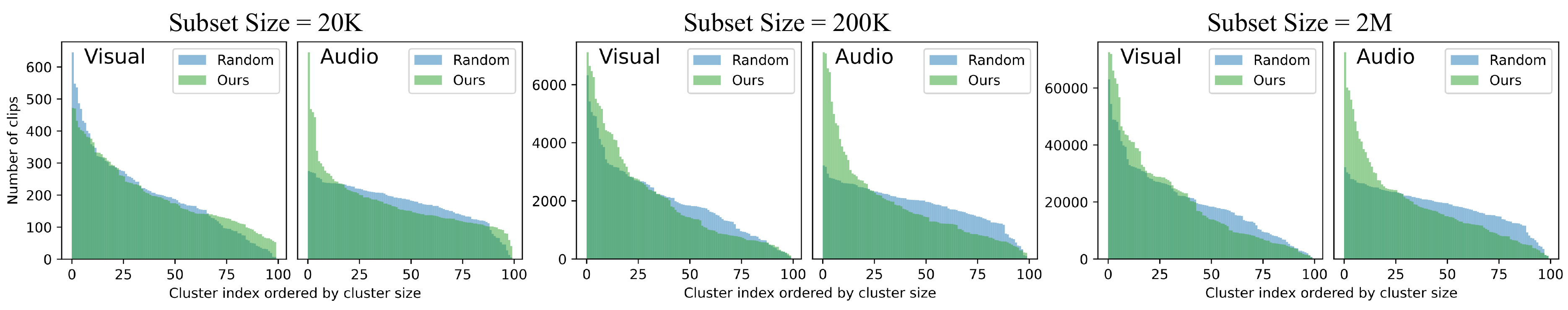}
    \caption{Histograms of cluster IDs from our curated subsets and randomly sampled subsets (with 100 cluster centroids). The blue histograms represent the case where samples are drawn uniformly random and thus is the unbiased representation of the concepts naturally appearing in the entire population.}
    \label{fig:histograms}
\end{figure*}

\section{On the Diversity of Concepts in Sampled Clips}
\label{sec:suppl_clip_diversity}

\subsection{Histogram of Cluster IDs}
To analyze the diversity of concepts contained in our curated dataset, we examine the histograms of cluster IDs from the chosen videos. Figure~\ref{fig:histograms} shows audio and visual histograms obtained from either our curated subsets or randomly sampled subsets at varying scales (20K, 200K, and 2M). To obtain these, we cluster the features from the last layer of audio and visual feature extractors, respectively, and plot the histograms of cluster IDs. For the purpose of visualization we sort the cluster indices by the cluster size in a decreasing order (and thus the cluster IDs do not match between ``Random'' and ``Ours'' in each of the plots). The histograms from random subsets represent the natural distribution of the entire video population.

In the visual domain, the curated datasets (green histograms) mostly follow the original cluster distributions (which is reflected in the blue histogram in each subplot). This indicates that the visual concept distribution largely follows the natural distribution in the entire population, suggesting that our subset contains visual concepts that are as diverse as the entire set.

On the other hand, the audio clusters show noticeable concentration in distribution after subset selection. Upon close inspection of videos from the largest audio clusters, we observe that our curated datasets tend to choose videos from clusters with high audio-visual correspondence (e.g., videos of a single person speaking with no other sound in background) while random sampling tend to choose videos from clusters with no apparent audio-visual correspondence (e.g., videos of multiple people taking with background music/noise). This shows that the concentration in the audio histograms is caused by filtering out videos of low audio-visual correspondence, which is a highly desirable artifact in the curated subset.

\subsection{Qualitative Analysis of Audio-Visual Clustering Results}
\begin{figure*}[!ht]
    \centering
    \includegraphics[width=0.97\textwidth]{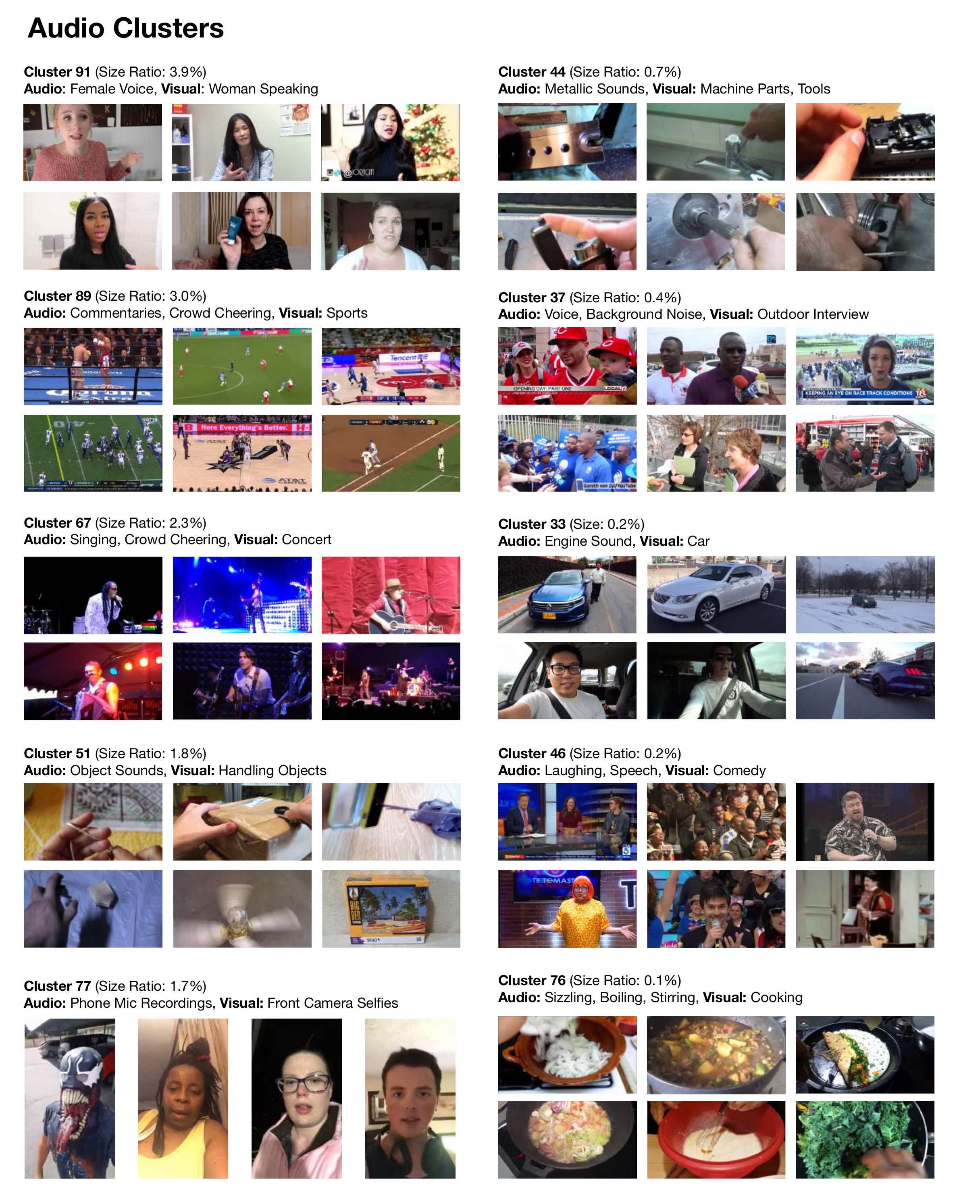}
    \caption{Representative samples and concepts derived from a manual inspection of 100 \textit{audio} clusters of the 2M subset. 
    We show samples from the five largest clusters on the left column and those from the five smallest clusters on the right.
    Each cluster captures distinctive audio-visual concepts, indicating that our curated subset contains various concepts with high audio-visual correspondence.}
    \label{fig:samples_per_clusters}
\end{figure*}

\begin{figure*}[!ht]
    \centering
    \includegraphics[width=0.97\textwidth]{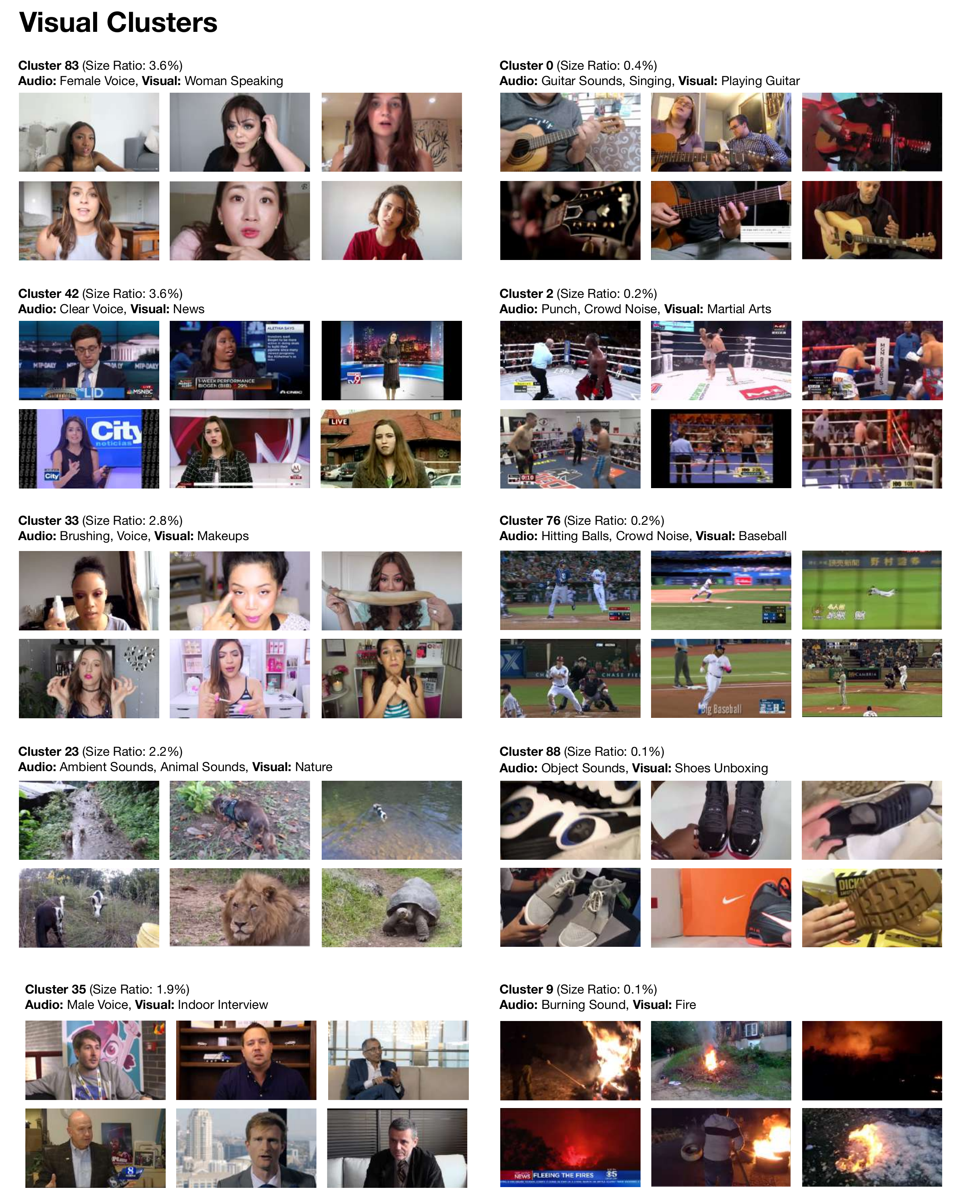}
    \caption{Representative samples and concepts derived from a manual inspection of 100 \textit{visual} clusters of the 2M subset. 
    We show samples from the five largest clusters on the left column and those from the five smallest clusters on the right.
    Each cluster captures distinctive audio-visual concepts, indicating that our curated subset contains various concepts with high audio-visual correspondence.}
    \label{fig:samples_per_clusters_visual}
\end{figure*}

To further investigate the diversity of concepts appearing in our subsets, we manually inspect audio and visual clustering results in the 2M dataset and compare the concepts appearing in the largest clusters to those in the smallest ones.
Figure~\ref{fig:samples_per_clusters} and Figure~\ref{fig:samples_per_clusters_visual} show representative videos from the five largest and five smallest clusters obtained from audio and visual clustering results, respectively.
Figure~\ref{fig:samples_per_clusters} (from audio clusters) suggests that our curated dataset contains diverse concepts including general sound categories (e.g., voice and objects sounds) as well as specific topics (e.g., outdoor interview and cooking).
Similarly, Figure~\ref{fig:samples_per_clusters_visual} (from visual clusters) also suggests that our dataset contains diverse concepts including both natural (e.g., animals and fire) and human sounds (e.g., makeup and playing guitar).
Clips from larger clusters (depicted in the left column of Figure~\ref{fig:samples_per_clusters} and Figure~\ref{fig:samples_per_clusters_visual}) contain clear and isolated sound sources, while sounds of smaller clusters (the right column) are less distinguishable due to multiple sound sources or background noise.
Our dataset also captures several audio-visual concepts that existing datasets (such as VGG-Sound~\cite{chen2020vggsound} and AudioSet~\cite{gemmeke2017audio}) do not offer.
For instance, in Figure~\ref{fig:samples_per_clusters}, the 77th cluster contains videos recorded from a front-facing camera with voice recordings from a phone mic, and the 46th cluster contains videos of comedians performing exaggerated body actions with the sound of crowd (cheering and laughter).
The 88th cluster in Figure~\ref{fig:samples_per_clusters_visual} contains shoes unboxing videos.


\section{Weighted Summation of Layer Scores (Section~\ref{ssec:correspondence_results})}
\label{sec:suppl_layer_weight}
Table~\ref{tab:weighted_layer_summation} compares different layer weighting schemes in clustering-based MI estimation, which shows that our multi-layer approach is generally robust to weight distributions. We explored two alternative weighting schemes: a \textit{linear($k$)} function with slope $k$ and an \textit{exp($k$)} function with slope $e^k$; we used \textit{uniform} weights in the main paper. We can see that precision is stable under a linear weighting scheme; the robustness comes from the \texttt{Combination} pairing approach which computes an average MI between all possible combinations across layers. However, precision drops significantly when the weights have a steep slope (e.g., \textit{exp(-10)}), which is a degenerate case similar to the single-layer approach reported in Table~\ref{tab:correspondence_similarity} of the main paper.

\begin{table}[!ht]
    \centering
    \small
    \setlength{\tabcolsep}{3.1pt}
    \begin{tabular}{l|c|c|c|c|c|c}
        \toprule
        \multirow{2}{*}{Method} & \multicolumn{5}{c|}{Layer Weights} & \multirow{2}{*}{Precision} \\
        \cline{2-6}
         & 1 & 2 & 3 & 4 & 5 & \\ \hline \hline
        exp(-10) & 5e+10 & 2e+04 & 1 & 5e-05 & 2e-09 & 50.791 \\
        exp(-1) & 7.4 & 2.7 & 1 & 0.4 & 0.1 & 65.374 \\
        exp(1) & 0.1 & 0.4 & 1 & 2.7 & 7.4 & 79.858 \\
        exp(10) & 2e-09 & 5e-05 & 1 & 2e+04 & 5e+10 & 57.880 \\ \hline
        linear(-0.50) & 1.9 & 1.5 & 1 & 0.5 & 0.1 & 88.018 \\
        linear(-0.25) & 1.5 & 1.2 & 1 & 0.8 & 0.5 & 88.673 \\
        linear(0.25) & 0.5 & 0.8 & 1 & 1.2 & 1.5 & 88.777 \\
        linear(0.50) & 0.1 & 0.5 & 1 & 1.5 & 1.9 & 87.997 \\ \hline
        \textbf{Uniform (Ours)} & 1 & 1 & 1 & 1 & 1 & 88.705 \\ \hline
    \end{tabular}
    \vspace{1em}
    \caption{Different layer weighting schemes in clustering-based MI estimation using Kinetics-Sounds with \texttt{Combination} pairing.}
    \label{tab:weighted_layer_summation}
\end{table}

\section{Details of Linear Evaluation on Downstream Tasks (Section~\ref{ssec:downstream})}
\label{sec:suppl_linear_eval}

\begin{table*}[!ht]
\centering
\small
\setlength{\tabcolsep}{3.1pt}
\begin{tabular}[t]{l|c|cc|cc|cc}
\toprule
\multirow{2}{*}{Size} & \multirow{2}{*}{Pretrain} & \multicolumn{2}{c|}{UCF101} & \multicolumn{2}{c|}{ESC-50} & \multicolumn{2}{c}{Kinetics-Sounds} \\
\cline{3-4}
\cline{5-6}
\cline{7-8}
& & top-1 & top-5 & top-1 & top-5 & top-1 & top-5 \\ \midrule
- & Random Init & 11.48 & 29.21 & 8.35 & 34.85 & 20.31 & 47.03 \\ \midrule
\multirow{4}{*}{20K} & Kinetics-Sounds & 33.51 & 64.47 & 49.40 & 81.85 & 49.98 & 82.15 \\
& Random Set & 36.34 & 66.59 & 46.95 & 79.30 & 45.19 & 77.25 \\
& \textbf{Clustering (Ours)} & 46.28 & 75.24 & 50.55 & 81.30 & 55.78 & 85.15 \\ \midrule
\multirow{4}{*}{200K} & VGG-Sound & 49.55 & 78.60 & 65.55 & 90.95 & 55.59 & 86.46 \\
& Random Set & 34.33 & 63.92 & 45.80 & 78.45 & 44.15 & 76.88 \\
& Contrastive & 45.10 & 76.46 & 56.90 & 85.00 & 53.80 & 85.26 \\
& \textbf{Clustering (Ours)} & 50.19 & 78.89 & 62.80 & 89.50 & 56.12 & 84.10 \\ \midrule
\multirow{4}{*}{2M} & AudioSet & 55.54 & 83.94 & 65.05 & 90.70 & 57.46 & 86.72 \\
& Random Set & 41.12 & 72.24 & 52.75 & 83.55 & 48.30 & 79.54 \\
& Contrastive & 45.87 & 75.80 & 58.85 & 87.10 & 53.68 & 83.05 \\
& \textbf{Clustering (Ours)} & 55.63 & 83.92 & 65.10 & 90.50 & 57.48 & 87.19 \\ \midrule
10M & \textbf{Clustering (Ours)} & 74.21 & 93.82 & 74.20 & 93.40 & 67.71 & 92.14 \\ \midrule
100M & \textbf{Clustering (Ours)} & 86.10 & 97.94 & 86.95 & 97.45 & 75.42 & 95.88 \\ \bottomrule
\end{tabular}
\vspace{1em}
\caption{Linear evaluation of representations pretrained on different datasets. We report the top-1/5 accuracies (\%) of video classification on UCF101~\cite{soomro2012ucf101}, audio classification on ESC-50~\cite{piczak2015dataset} and audio-visual classification on Kinetics-Sounds~\cite{arandjelovic2017look}. We average the accuracies across the official splits of UCF101 (three splits) and ESC-50 (five splits).}
\label{tab:downstream_results}
\end{table*}

Table~\ref{tab:downstream_results} shows the results of liner evaluation on downstream tasks, which were also shown in the bar chart of the main paper, Figure~\ref{fig:linear evaluation}; we reproduced here to compensate for the lack of readability of the bar chart.

\subsection{Experimental Settings}
We pretrain audio-visual models in a contrastive manner~\cite{chen2020simple} on different datasets. Specifically, we attach MLP projection heads on top of audio and visual feature extractors, respectively, and train the whole model end-to-end using the noise-contrastive loss (see Eqn.~\ref{eq:contrastive} of the main paper).
As for the visual and audio backbone feature extractors, we use 3D ResNet-50~\cite{carreira2017quo} and ResNet-50~\cite{he2016deep}, respectively.
Each of the MLP projection head is composed of two fully-connected layers with ReLU~\cite{nair2010icml} activation, and produces the embeddings of dimension 128.
We pretrain the model for 50 epochs with a batch size 64.
We use the AMSGrad variant~\cite{reddi2018convergence} of AdamW~\cite{loshchilov2017decoupled} optimizer with a learning rate 1e-3, $\beta_1=0.9$, $\beta_2=0.999$ and an L2 weight decay of 1e-5.
We apply learning rate warm-up for the first 20,000 iterations followed by a linear decay of learning rate.

For linear evaluation on downstream tasks, we attach a linear classifier on top of the pretrained feature extractors and train it from scratch while  fixing the parameters of the feature extractors. We use only the visual CNN for action recognition on UCF101~\cite{soomro2012ucf101} and only the audio CNN for sound classification on ESC50~\cite{piczak2015dataset}. For audio-visual action recognition on Kinetics-Sounds~\cite{arandjelovic2017look}, we concatenate audio-visual features before feeding them as input to the linear classifier.
We apply dropout~\cite{hinton2012improving} with a 50\% rate before the linear classifier.
We train the model for 30 epochs with a batch size of 1024 on ESC-50~\cite{piczak2015dataset}, for 10 epochs with a batch size of 64 on UCF101~\cite{soomro2012ucf101} and for 5 epochs with a batch size of 64 on Kinetics-Sounds.
We use the Adam~\cite{kingma2015adam} optimizer with a learning rate 1e-2, $\beta_1=0.9$, $\beta_2=0.999$ and an L2 weight decay of 5e-6.

\subsection{Impact of the Number of Centroids}
\begin{figure*}[!h]
   \centering
   \includegraphics[width=\textwidth]{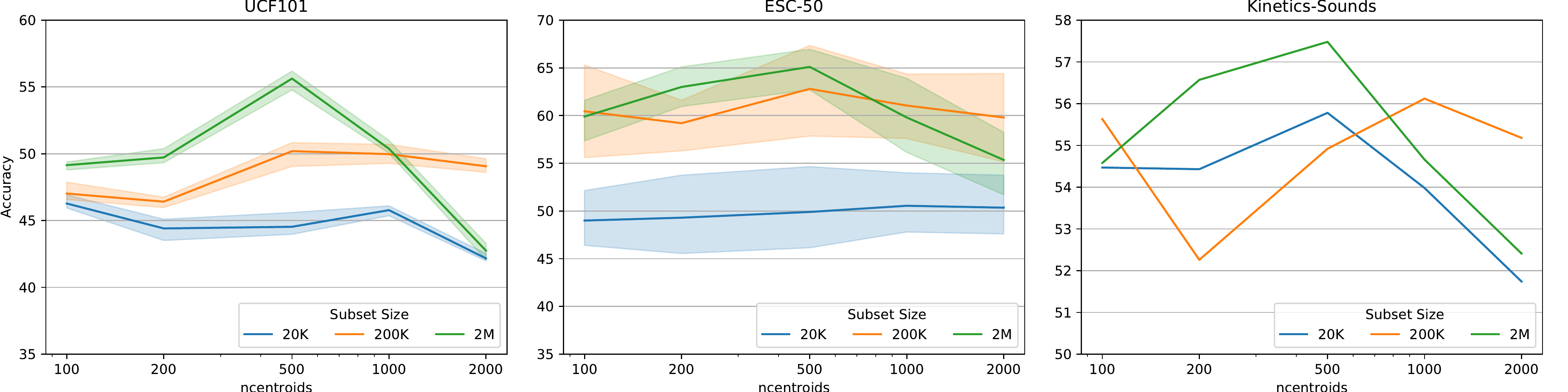} 
   \caption{Linear evaluation of representations pretrained on the datasets that are constructed by our clustering-based approach. We report the top-1 accuracy (\%) on UCF101~\cite{soomro2012ucf101}, ESC-50~\cite{piczak2015dataset}, and Kinetics-Sounds~\cite{arandjelovic2017look}, grouped by the number of cluster centroids. The shaded regions show 99\% confidence intervals obtained by runs over the official splits of UCF101 (3 splits) and ESC-50 (5 splits).}
   \label{fig:linear_eval_lines_centroids}
\end{figure*}

To visualize the impact of the number of clusters in our clustering-based approach, we group the results by the number of clusters as shown in Figure~\ref{fig:linear_eval_lines_centroids}.
Notice that the number of clusters is not positively correlated with downstream task performance.
Instead, clustering with about 500 clusters seems to yield the best performance.
Also, experiments using the largest number of centroids ($C=2000$) show low accuracy consistently across all datasets and subset sizes.
This confirms our findings in Section~\ref{ssec:correspondence_results} of the main paper: over-clustering tends to have a negative impact on the quality of the selected subset.
We believe that this happens because, as the number of clusters increases,
samples with homogeneous concepts in large clusters are scattered into small clusters sharing similar concepts.
When we do not have many references to compare as in the early stage of subset selection, this fragmentation effect inhibits sample count sharing between conceptually similar small clusters, complicating the clustering-based MI estimation.

\section{More Discussion on Subset Selection (Section~\ref{ssec:selection_clustering})}
\label{sec:suppl_subset_selection}

\subsection{Greedy Algorithm}
\label{ssec:suppl_greedy}

We provide the details of the greedy algorithm~\cite{nemhauser1978analysis} that is approximated using the batch greedy algorithm~\cite{chen2013near}.
As shown in Algorithm~\ref{alg:greedy}, the greedy algorithm needs to re-evaluate the clustering-based MI estimator $F$ on all the remaining candidates in each iteration.
Thus, the time complexity is $O(N^2)$ where $N$ is the size of the initial dataset $\mathbf{D}$.

On the other hand, the batch greedy algorithm approximates this by selecting the next element to be included in the solution within only a randomly chosen batch, not the entire candidates.
This is shown in Algorithm~\ref{alg:batch_greedy_supp} below (same as Algorithm~\ref{alg:batch_greedy} of the main paper; reproduced here for easy comparison).

\subsection{Batch Greedy Subset Selection}
\label{ssec:suppl_batch_greedy}
\begin{figure*}[!h]
    \centering
    \includegraphics[width=\textwidth]{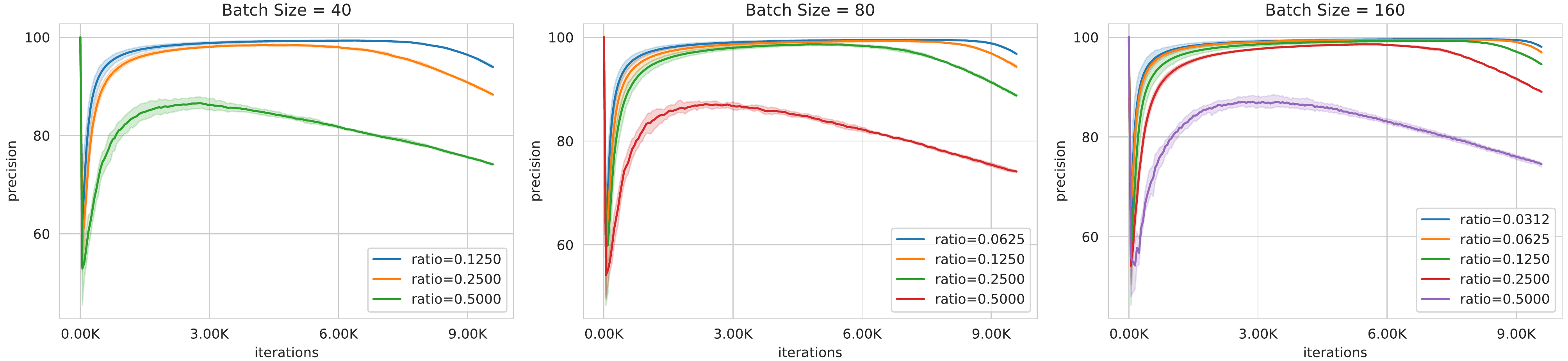}
    \caption{Precisions of Batch greedy algorithm with varying ratios of selection size to batch size, $s/b$ (x axis: iterations, y axis: precision). We group the plots by the batch size: $b=40, 80, 160$ from left to right. The shaded regions show 99\% confidence intervals obtained by five runs on Kinetics-Sounds. The batch greedy algorithm is robust when the ratio is $\leqslant$ 25\%, regardless of the batch size.}
    \label{fig:suppl_batch_greedy}
\end{figure*}

When using the batch greedy algorithm for subset selection, the batch size $b$ and the selection size $s$ affect the quality of the selected subsets.
We explore various $(b, s)$ configurations on Kinetics-Sounds~\cite{arandjelovic2017look}, as shown in Figure~\ref{fig:suppl_batch_greedy}.
Note that the performance gap between different batch sizes is small. The precision 93.9\%, 94.3\% and 94.6\% are respectively obtained when using batch sizes $b=40, 80, 160$ with the same ratio of selection size to batch size $s/b=12.5\%$.
On the contrary, the value of $s/b$ highly affects the retrieval performance across all the batch sizes examined; the performance drops sharply as the ratio exceeds 25\% regardless of the batch size.
As stated in Section~\ref{ssec:correspondence_results} of the main paper, we construct the dataset to have an equal number of positive and negative pairs and the drop in robustness manifests itself when the selection ratio $s/b$ exceeds the \textit{easy positive} ratio of 25\%.

\section{Details of Automatic Dataset Curation}
\label{sec:suppl_hyperparams}

Here, we describe the details of subset selection via (i) NCE-based MI estimation and (ii) clustering-based MI estimation. To construct datasets, we vary scales of 20K, 200K and 2M. Based on the results at the three scales, we also generate a version with 10M videos using the clustering-based approach.

\begin{algorithm}[h]
\SetAlgoNoLine
\textbf{Input:} initial dataset $\mathbf{D}$, clustering-based MI estimator $F$, target subset size $M$\\
\textbf{Output:} $\mathbf{X} \subseteq \mathbf{D}, |\mathbf{X}| = M$\\
$\mathbf{X}_0 \gets \emptyset$\\
\For{$i = 0$ \KwTo $M-1$}{
    $x \gets \argmax_{x \in \mathbf{D} \backslash \mathbf{X}_i} F(\mathbf{X}_i \cup \{x\}) $\\
    $\mathbf{X}_{i+1} \gets \mathbf{X}_i \cup \{x\}$\\
}
$\mathbf{X} \gets \mathbf{X}_M$\\
\textbf{Return} $\mathbf{X}$
\caption{Greedy Algorithm}
\label{alg:greedy}
\end{algorithm}

\begin{algorithm}[!h]
\SetAlgoNoLine
\DontPrintSemicolon
\textbf{Input:} initial dataset $\mathbf{D}$, clustering-based MI estimator $F$, target subset size $M$, batch size $b$, selection size $s$\\
\textbf{Output:} $\mathbf{X} \subseteq \mathbf{D}, |\mathbf{X}| = M$\\
$\mathbf{X}_0 \gets \emptyset, i \gets 0$\\
\While{$|X_i| < M$}{
    Randomly sample $\mathbf{B} \subseteq \mathbf{D} \backslash \mathbf{X}_{i}, |\mathbf{B}| = b$\\
    $\mathbf{Y}_0 \gets \emptyset, j \gets 0$\\
    \While{$j < s$}{
        $x \gets \argmax_{x \in \mathbf{B} \backslash \mathbf{Y}_j} F(\mathbf{X}_i \cup \mathbf{Y}_{j} \cup \{x\})$\\
        $\mathbf{Y}_{j+1} \gets \mathbf{Y}_{j} \cup \{x\}, j \gets j + 1$\\
        \lIf{$|\mathbf{X}_i \cup \mathbf{Y}_j| = M$}{
            break
        }
    }
    $\mathbf{X}_{i+1} \gets \mathbf{X}_i \cup \mathbf{Y}_j, i \gets i + 1$
}
$\mathbf{X} \gets \mathbf{X}_i$\\
\textbf{Return} $\mathbf{X}$
\caption{Batch Greedy Algorithm (reproduced from the main paper for easy comparison)}
\label{alg:batch_greedy_supp}
\end{algorithm}
\noindent

\subsection{NCE-Based MI Estimation}
\label{ssec:suppl_nce}
We use the linear projection heads that transform audio and visual features into 128-dimension embeddings.
We randomly sample a subset of 100M clips from the initial 300M set that we crawl, and train on the subset for three epochs with a batch size $N_b=1,024$.
We use the AMSGrad variant of Adam optimizer~\cite{reddi2018convergence} with a learning rate 2e-4, $\beta_1=0.9$ and $\beta_2=0.999$.
We apply learning rate warm-up for the first 3 epochs followed by a linear decay of learning rate.

\subsection{Clustering-Based MI Estimation}
\label{ssec:suppl_clustering}

For SGD K-Means clustering, we train the cluster centroids with a mini-batch of size 100K for 100 epochs using a learning rate $\lambda=\textrm{1e-2}$.
When applying the batch greedy algorithm, we use the fixed batch size $b=10,000$ and the selection size $s=500$ (with a ratio of $s/b=0.05$), but vary the number of clusters $C \in \{100,200,500,1000,2000\}$ for each size of the datasets, except the dataset of 10M scale (we generate the dataset only with $C=500$ for computational reasons).

\section{Human Evaluation Interface (Section~\ref{ssec:human_evaluation})}
\label{sec:suppl_human_eval_interface}

\begin{figure*}[!h]
   \centering
   \includegraphics[width=0.97\textwidth]{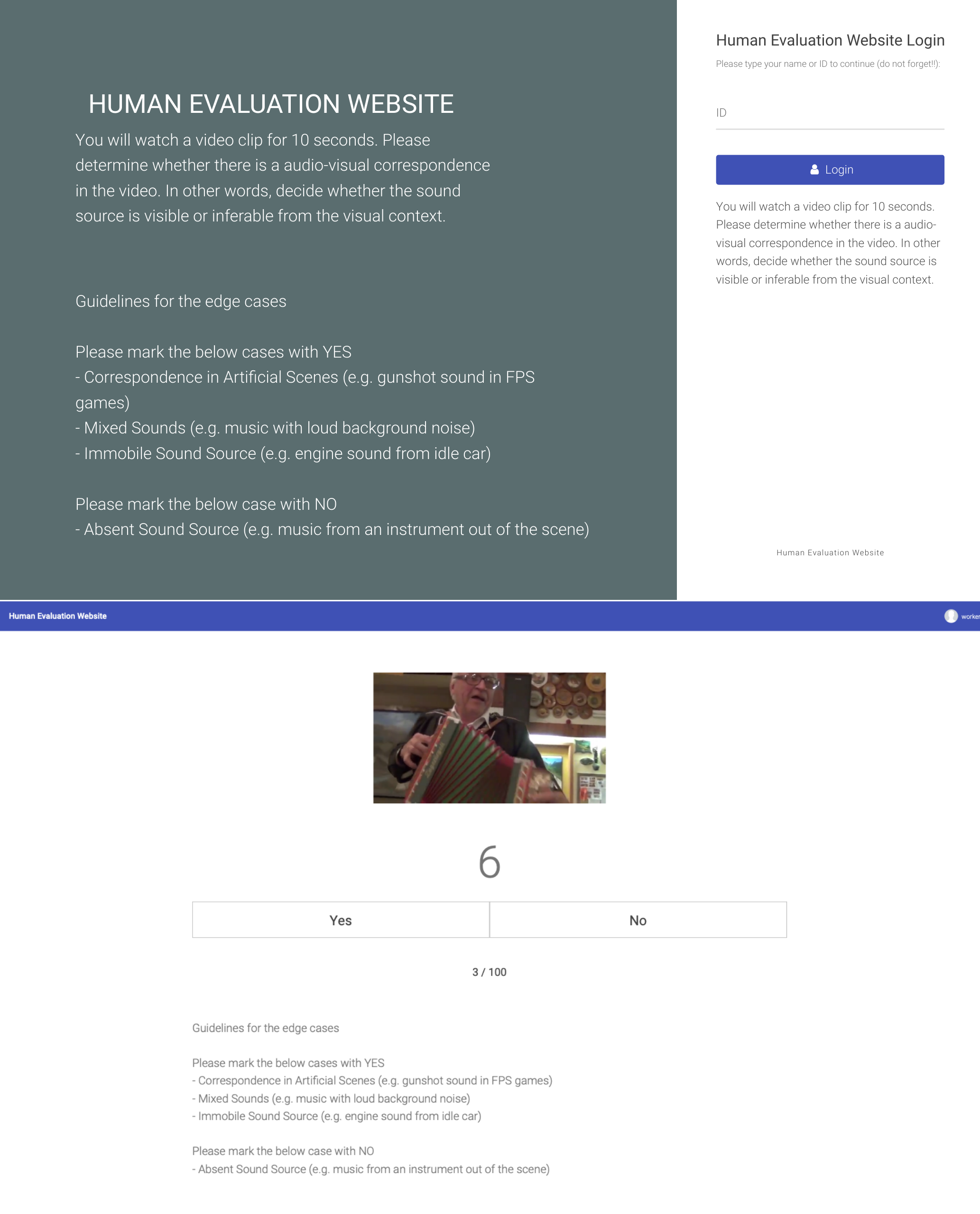} 
   \caption{Screenshots of the human evaluation interface. The introduction page (top) provides instructions to the annotators, and the test page (bottom) shows clips to the raters and receives the corresponding Yes/No responses.}
   \label{fig:human_eval_interface}
\end{figure*}
Figure~\ref{fig:human_eval_interface} shows the user interface we developed for human evaluation. We provide guidelines on how to assess audio-visual correspondence:

\begin{displayquote}
\textit{
You will watch a video clip for 10 seconds. Please determine whether there is audio-visual correspondence in the video. In other words, decide whether the sound source is visible or can be inferred from visual context.
} 
\end{displayquote}

After a pilot study we gathered feedback from experts and added additional guidelines to help disambiguate common edge scenarios (shown in Figure~\ref{fig:human_eval_interface}). 
Annotators are given one 10-second clip at a time and asked to provide a Yes/No answer judging whether or not there is audio-visual correspondence in the given clip. We do not provide a replay interface to collect intuitive response from the raters.

\clearpage

{\small
\bibliographystyle{ieee_fullname}
\bibliography{refs}
}

\end{document}